\documentclass[sigconf]{acmart}

\copyrightyear{2023}
\acmYear{2023}
\setcopyright{acmlicensed}
\acmConference[CIKM '23] {Proceedings of the 32nd ACM International Conference on Information and Knowledge Management}{October 21--25, 2023}{Birmingham, United Kingdom.}
\acmBooktitle{Proceedings of the 32nd ACM International Conference on Information and Knowledge Management (CIKM '23), October 21--25, 2023, Birmingham, United Kingdom}
\acmPrice{15.00}
\acmISBN{979-8-4007-0124-5/23/10}
\acmDOI{10.1145/3583780.3614759}

\usepackage{microtype}
\usepackage{graphicx}
\usepackage{subfigure}
\usepackage{booktabs} 

\usepackage{hyperref}

\newcommand{\xv}{\boldsymbol{x}}
\newcommand{\hv}{\boldsymbol{h}}
\newcommand{\gv}{\boldsymbol{g}}
\newcommand{\uv}{\boldsymbol{\mu}}

\usepackage{amsmath}
\usepackage{amssymb}
\usepackage{mathtools}
\usepackage{amsthm}
\usepackage{multirow}
\usepackage[capitalize,noabbrev]{cleveref}
\usepackage{bbm}
\theoremstyle{plain}

\theoremstyle{definition}

\theoremstyle{remark}

\usepackage[textsize=tiny]{todonotes}
\newcommand{\eat}[1]{}
\usepackage[ruled,linesnumbered]{algorithm2e}

\begin{document}

\title{A Co-training Approach for Noisy Time Series Learning}

\author{Weiqi Zhang}
\authornote{This work was done during the internship at Huawei Noah’s Ark Lab.}
\affiliation{%
  \institution{The Hong Kong University of Science and Technology}
  \city{Hong Kong SAR}
  \country{China}}
\email{wzhangcd@connect.ust.hk}

\author{Jianfeng Zhang}
\affiliation{%
  \institution{Huawei Noah’s Ark Lab}
  \city{Shenzhen}
  \country{China}}
\email{zhangjianfeng3@huawei.com}

\author{Jia Li}
\authornote{Corresponding Author}
\affiliation{%
  \institution{The Hong Kong University of Science and Technology (Guangzhou)}
  \city{Guangzhou}
  \country{China}}
\affiliation{%
  \institution{The Hong Kong University of Science and Technology}
  \city{Hong Kong SAR}
  \country{China}}
\email{jialee@ust.hk}

\author{Fugee Tsung}
\affiliation{%
  \institution{The Hong Kong University of Science and Technology (Guangzhou)}
  \city{Guangzhou}
  \country{China}}
\affiliation{%
  \institution{The Hong Kong University of Science and Technology}
  \city{Hong Kong SAR}
  \country{China}}
\email{season@ust.hk}

\begin{abstract}

In this work, we focus on robust time series representation learning. Our assumption is that real-world time series is noisy and complementary information from different views of the same time series plays an important role while analyzing noisy input. Based on this, we create two views for the input time series through two different encoders. We conduct co-training based contrastive learning iteratively to learn the encoders. 
Our experiments demonstrate that this co-training approach leads to a significant improvement in performance. Especially, by leveraging the complementary information from different views, our proposed TS-CoT method can mitigate the impact of data noise and corruption. Empirical evaluations on four time series benchmarks in unsupervised and semi-supervised settings reveal that TS-CoT outperforms existing methods. Furthermore, the representations learned by TS-CoT can transfer well to downstream tasks through fine-tuning\footnote{ Our code is available at \url{https://github.com/Vicky-51/TS-CoT}.}.
\end{abstract}

\ccsdesc[500]{Information systems~Data mining}
\ccsdesc[500]{Computing methodologies~Learning latent representations}
\ccsdesc[500]{Mathematics of computing~Time series analysis}
\keywords{Time Series; Co-training; Contrastive Learning; Noisy Data}

\maketitle

\vspace{-2mm}
\section{Introduction}

Time series data is ubiquitous in the real world \cite{cryer1986time, ching2018opportunities, ye2022learning} and the analyses over it are fundamental for lots of applications, especially in healthcare \cite{stevner2019discovery, li2018tatc, saab2020weak, li2022towards,Zhang_2023}, finance \cite{chan2004time, franses2000non, abilasha2022deep, du2021adarnn}, traffic \cite{han2021dynamic, cirstea2022towards, shao2022pre, li2019predicting, kim2022residual}, electricity \cite{deng2021st, ye2022learning, wu2023timesnet}, and industrial production areas \cite{carbonneau2008application, feng2021time}. However, the accessibility of ground-truth labels is very time-consuming and expensive in practice, making it challenging to directly apply classical supervised learning methods \cite{zhang2022tfad}. Additionally, collected time series data is prone to noise and corruption, which presents new challenges. In this vein, there is an urgent call for effective and robust unsupervised and semi-supervised time series representation learning methods in various time series analysis tasks \cite{tonekaboni2020unsupervised}.

\begin{figure}[tb]
  \centering
  \includegraphics[width=0.95\columnwidth]{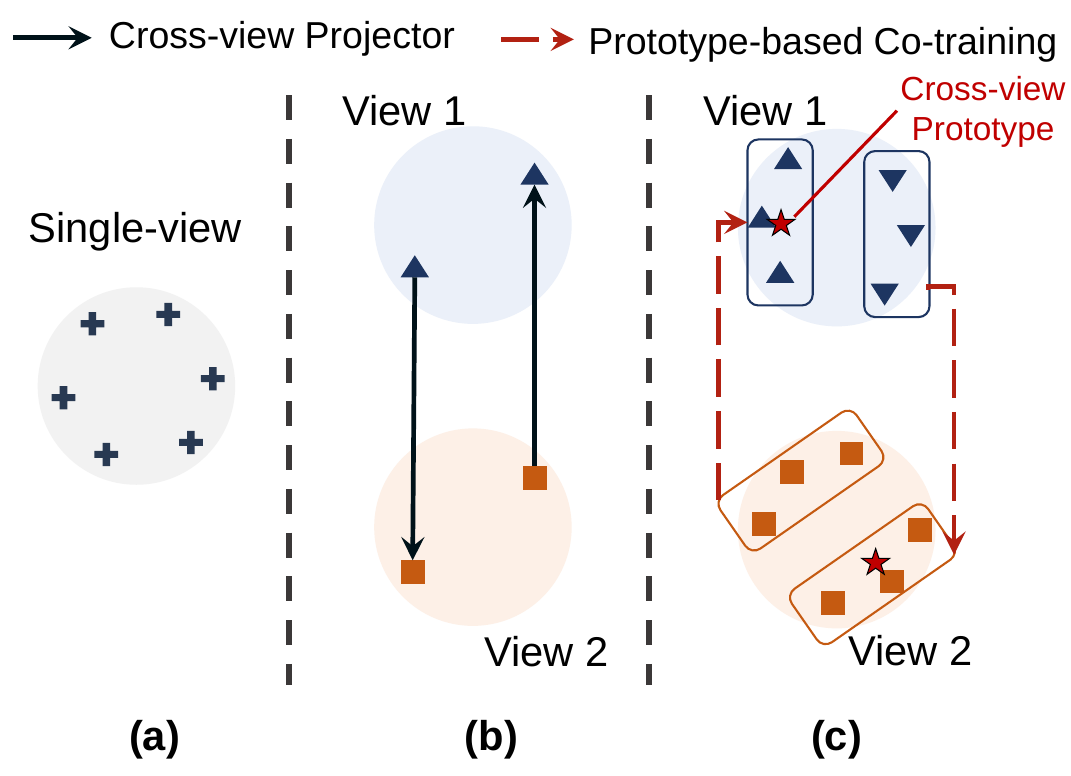}
  \vspace{-5mm}
  \caption{Different paradigms for time series contrastive learning in single-view and multi-view settings. The circle denotes the unit hypersphere of the learned representations. (a): Single-view contrastive learning method shows alignment and uniformity of feature distributions \cite{wang2020understanding}. (b): Time-Frequency consistency learning algorithm proposed in \cite{zhang2022self}. (c): Prototype-based co-training algorithm proposed in this work.}
  \label{figure:intro}
  \vspace{-5mm}
\end{figure}

Time series representation learning aims to extract informative embedding vectors from the original time series input for downstream tasks. As a powerful unsupervised representation learning framework, contrastive learning has been adopted by many existing works. In this learning paradigm, positive and negative pairs are obtained by data augmentation and pairing with other instances in the same mini-batch, respectively \cite{yue2022ts2vec, eldele2021time, woo2022cost}. For time series contrastive learning, there exist two paradigms: single-view and multi-view, as shown in Figure \ref{figure:intro}(a-b). 

For the single-view learning paradigm, though uniformity and alignment of feature distribution could be observed, it is hard to capture the macro semantic structure of the hidden representation space, which has received much attention in recent studies \cite{haochen2021provable, li2020prototypical, wang2021pico}. Besides, the representation learning performance would heavily rely on the size of the mini-batch, which is crucial to construct the negative pairs. On the other hand, existing multi-view learning approaches mainly focus on finding nearest neighbors as positive pairs \cite{li20213d, han2020self}, which will lead to heavy computational costs for storing the embeddings of negative samples. For time series representation learning, \citet{zhang2022self} encourage the time and frequency signals to have exactly similar embeddings after a linear projector (as shown in Figure \ref{figure:intro}(b)), which may be \eat{easily violated} difficult to achieve in practice because of the complementary and disagreed information of different views.

\begin{figure}[tb]
  \centering
  \includegraphics[width=0.9\columnwidth]{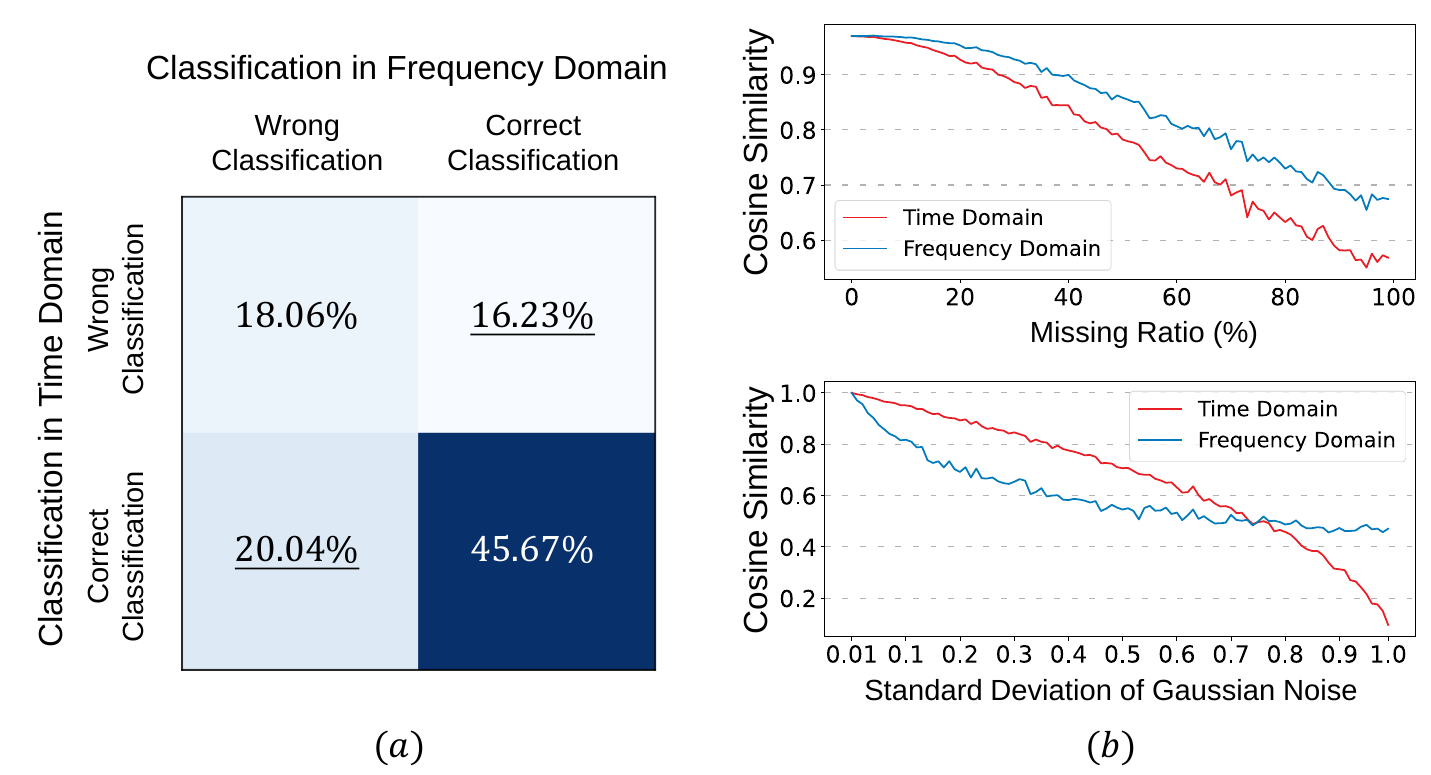}
  \vspace{-5mm}
  \caption{(a): Single-view classification results on Waveform dataset. (b): Cosine similarity between the noisy time series and noiseless signals across different views.}
  \label{figure:intro_example}
  \vspace{-6mm}
\end{figure}

Significantly, based on our experimental analysis (as depicted in Figure \ref{figure:intro_example}), complementary information from different views plays an important role while analyzing time series data. First, when conducting vanilla linear classification solely on single-view data, we encounter numerous samples that are misclassified in one view but correctly classified in another (Figure \ref{figure:intro_example}(a)). This indicates the presence of complementary information in multi-view time series data. Furthermore, the influence of noisy data could be alleviated by exploring multi-view signals. Empirically, we simulate noisy signals by introducing random missing values and Gaussian noise. We compare the cosine similarity between the noisy time series and noiseless signals across different views (Figure \ref{figure:intro_example}(b)). Our findings reveal that, in some cases, temporal signals exhibit greater resilience to noise, while in other cases, spectral signals perform better. Consequently, utilizing the multi-view complementary information can significantly enhance model robustness against noise.

Moreover, existing time series time-frequency consistency learning algorithms, such as the one illustrated in Figure \ref{figure:intro}(b), only focus on the local structure of each instance and ignore the global structure of the whole hidden representation space. In this work, we hope to achieve the consistency of different views globally by maximizing the agreement of the semantic structures in their hidden spaces. Motivated by the success of co-training in unsupervised and semi-supervised learning \cite{nigam2000analyzing, kumar2011co}, we design a time series co-training algorithm (TS-CoT) for better structure discovery by incorporating cross-view information and calibrating semantic differences.

To be specific, as shown in Figure \ref{figure:intro}(c), we assume the underlying hidden space is multimodal, and denote the center of each modal as a prototype. We propose to maintain two sets of prototypes to describe the hidden structures of two views during co-training. Each representation will be pushed towards its cross-view prototype, which is the center of the points belonging to the same prototype in the other view. By involving the cross-view prototypes, which is a global embedding vector, the limitation of batch size will be alleviated. In each iteration, the sample can not only perform contrastive learning with other samples in the same mini-batch but can also access the whole hidden space through cross-view prototypes. Meanwhile, compared with nearest neighbors-based multi-view representation learning approaches, our proposed prototype-based method can effectively reduce the memory cost as it does not require storing the negative pairs.

In addition, we propose a moving-average prototype updating strategy during co-training, which further declines the computational cost and alleviates the instability of unsupervised clustering algorithms compared with purely clustering-based prototypical contrastive learning \cite{li2020prototypical}. Our proposed TS-CoT can be applied in both unsupervised and semi-supervised tasks and the extensive experiments show that TS-CoT has superior performance. The contributions of this paper can be summarized as follows:
\begin{itemize}
    \item We propose TS-CoT, a novel and general framework for time series representation learning in multi-view settings. The proposed TS-CoT can be extended to both unsupervised and semi-supervised tasks, where it works consistently well with sparsely labeled or entirely no labels.
    \item We investigate the robustness by exploring the complementary information of different views. To the best of our knowledge, TS-CoT is the first proposed representation learning method for noisy and corrupted time series data.

    \item TS-CoT shows superior empirical performance by extensive experiments. Significant improvements could be observed in different downstream tasks. Ablation study, robustness analysis, and sensitivity analysis further demonstrate the effectiveness of our proposed method.
\end{itemize}

\vspace{-2mm}
\section{Related Work}
\subsection{Time Series Representation Learning}
Time series representation learning methods, whose goal is to encode the original time series into the latent representation space and obtain an informative embedding, have been rapidly developed in recent years. Existing works could be further divided into Transformer based and contrastive learning based representation learning methods. Transformer based methods learn the vector representations of target time series in a ``denoising" way \cite{zerveas2021transformer}. Furthermore, in \cite{chowdhury2022tarnet}, \citeauthor{chowdhury2022tarnet} propose a task-aware transformer-based model to facilitate downstream tasks. Contrastive based methods aim at maximizing the agreement of positive pairs and pulling away the negative pairs. In \cite{oord2018representation}, a contrastive predictive coding framework is proposed by treating the data of the same series as positive samples and optimizing the InfoNCE. TNC \cite{tonekaboni2020unsupervised} further improves the negative sampling strategy based on positive unlabeled learning. Meanwhile, TS-TCC \cite{eldele2021time}, TS2Vec \cite{yue2022ts2vec}, CoST \cite{woo2022cost}, Mixing-Up \cite{wickstrom2022mixing}, TF-C \cite{zhang2022self}, BTSF \cite{yang2022unsupervised} construct the positive sample pairs by conducting augmentations or combining different augmentations on time series data. \citet{eldele2021time} involve auto-regressive prediction to combine a strong augmentation and a weak augmentation. \citet{yue2022ts2vec} deploy hierarchical contrastive learning for two cropped subsequences of an input series.

Recently, several methods have considered both the original series and the spectral signals while modeling. For example, CoST \cite{woo2022cost} performs Fast Fourier Transform (FFT) on the latent embedding and obtains the representations in the spectral domain for contrastive learning. The design of CoST is based on an additional seasonal-trend disentangling assumption. BTSF \cite{yang2022unsupervised} relies on iterative matrix multiplication and factorization for spectral-temporal fusion. TF-C \cite{zhang2022self} turns to cross-space projectors to learn consistency between the time and frequency embedding of the same instance. However, existing works suffer from heavy computational or memory costs while learning the cross-view representations. Additionally, only instance-level contrastive learning is considered, which ignores the global semantic structure of the whole representation space and makes the size of the training mini-batch a crucial bottleneck for efficient implementation.

In addition, while traditional signal denoising has attracted much attention \cite{frusque2022robust, bayer2019iterative}, only a limited number of studies have focused on investigating the robustness of deep learning-based models against noisy time series input. \citet{zhang2023robust} apply localized stochastic sensitivity to a robust RNN model for better time series forecasting. In \cite{yoon2022robust}, \citet{yoon2022robust} extend the randomized smoothing strategy to establish a robust probabilistic time series forecasting framework. However, to our best knowledge, we are the first study for robust time 
 series self-supervised representation learning with noisy input.

\vspace{-3mm}
\subsection{Co-training Algorithms}
Co-training algorithm is a well-known semi-supervised learning paradigm that leverages multiple distinct views of the data \cite{blum1998combining, dasgupta2001pac}.  The theoretical explanations of co-training have also been discussed \cite{dasgupta2001pac, nigam2000analyzing, wang2007analyzing}. The key idea of co-training is to exploit complementary information from different views. In the classical form, two classifiers are trained separately on two sufficient and redundant views. Each learnable classifier labels some unlabeled data for the other. It has been proved that based on sufficient and redundant views, the co-training algorithm could decrease the generalization errors by maximizing different views' agreement over the unlabeled data. Except for the semi-supervised learning, co-training has also been extended for unsupervised multi-view clustering \cite{nigam2000analyzing, kumar2011co}. In \cite{kumar2011co}, \citeauthor{kumar2011co} incorporate the co-training strategy by integrating it with the unsupervised learning algorithm, spectral clustering.  Based on the assumption that one point would be assigned to the same clustering regardless of the view, the co-training based multi-view clustering algorithms empirically improve the performance and obtain the clusters agreed across different views.

There exists some work about co-training paired networks designed for contrastive representation learning, especially in the field of computer vision. A popular paradigm to perform cross-view consistency learning is to provide pseudo labels (e.g., by using k-nearest neighbors) for establishing the positive and negative samples in another view \cite{han2020self,li20213d}. Based on the pseudo-positive set, the self-supervised representation learning could be converted to a supervised one, where the UberNCE takes the place of the InfoNCE \cite{han2020self}. However, in this paradigm, only several nearest samples could be utilized for further contrastive learning. The high-level global structure of the representation space is hard to describe in this way. Meanwhile, the performance may be limited by the size of the memory bank, which stores part of the negative embeddings.

\begin{figure*}[tb]
  \centering
  \includegraphics[width=\textwidth]{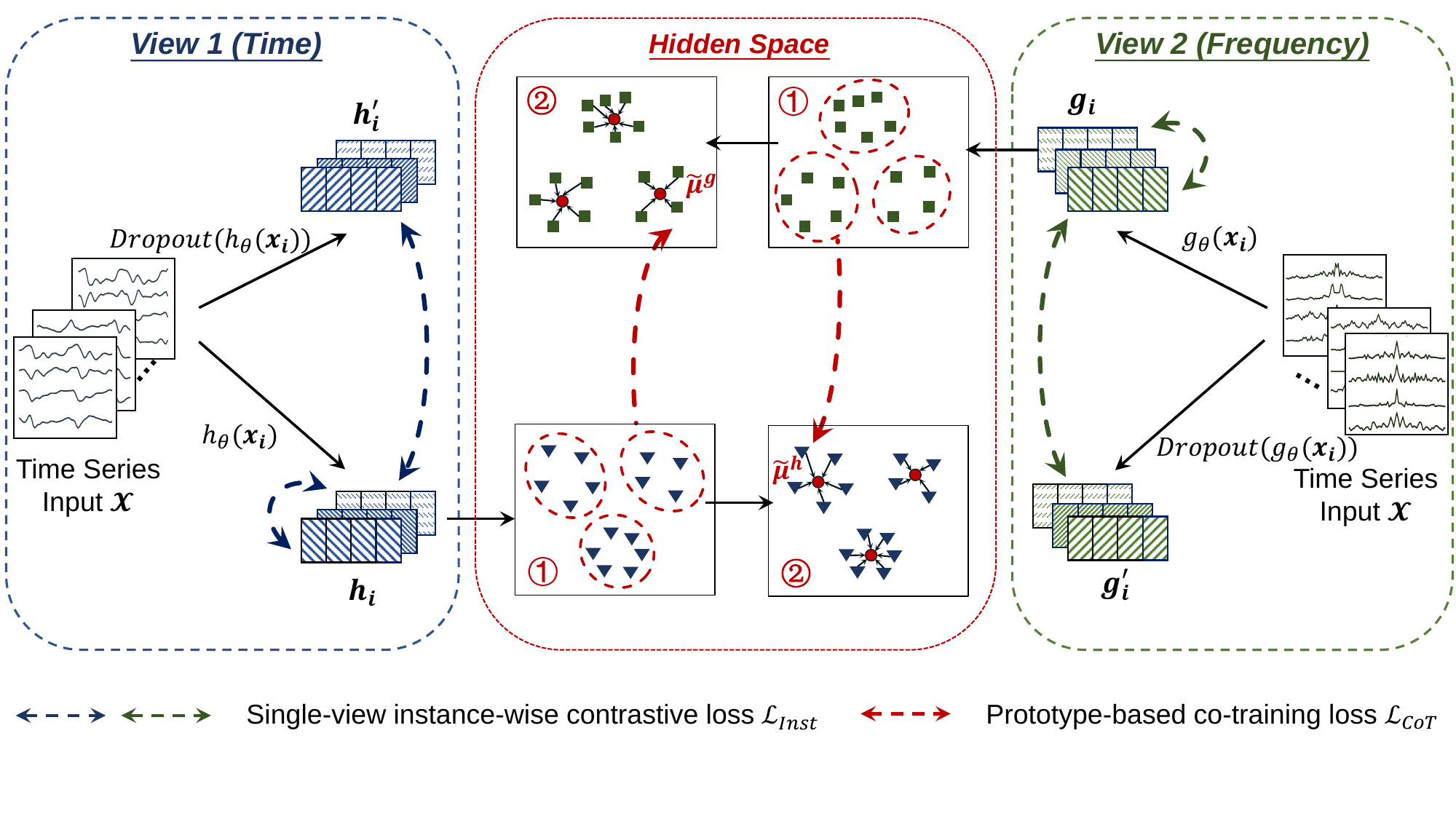}
  \vspace{-8mm}
  \caption{The framework of time series prototype-based co-training model. For illustration, we utilize the time and frequency domain as two views here. We first obtain the representations from the single-view encoders, where the instance-wise contrastive loss could be calculated. In the hidden space, the prototype-based co-training works in two steps. In step 1, we calculate the intra-view prototype assignments. In step 2, cross-view prototypes could be produced, based on which we calculate the prototype-based co-training loss by pushing the instances closer to their corresponding cross-view prototypes.}
  \label{figure:overall}
\end{figure*}

\section{Method}

\subsection{Problem Definition}

Let $\mathcal{X}=\left\{\xv_1, \xv_2, \ldots, \xv_N\right\}$ be a set of time series, where $\boldsymbol{x}_i \in \mathbb{R}^{T \times d}  $ is a $d$-dimension multivariate time series with length $T$. In our multi-view setting, we create two views for time series $\boldsymbol{x}_i$ through two learnable representation functions $h_{\theta}(\xv_i)$ and $g_{\theta}(\xv_i)$, which map the original time series into two latent spaces. The goal of our time series representation learning is to learn $h_{\theta}(\xv_i)$ and $g_{\theta}(\xv_i)$ through a co-training strategy. To simplify the notation, in following sections we denote $\hv_i=h_{\theta}(\xv_i)$ and $\gv_i=g_{\theta}(\xv_i)$, and two views are indicated by $h$ and $g$.

\subsection{Single-View Contrastive Learning}

To set up the co-training, two single-view encoders are first developed to map the input data into two latent spaces. Following the recent advances in contrastive learning, within every single view, the encoder learns the representations to align two augmented signals of the original instance. The effective and suitable time series augmentation selection is still a challenging task \cite{wen2020time}, and it is not the objective of this work. Thus, instead of conducting complicated transformations on the original time series, we adopt a straightforward dropout-based augmentation strategy in our implementation. As a widely-used technique in deep neural networks, the dropout operation does not change the semantic information, thus it is a natural way to augment the data for contrastive learning. More precisely, for each input $\xv$, we perform random dropout in every intermediate embedding layer within the neural network $h_{\theta}$ (resp. $g_{\theta}$) to create an augmentation for $\xv$, i.e.,
$\hv'_{i}=\operatorname{Dropout}(h_{\theta}(\xv_{i})).$

We define the $(\hv_i, \hv_i')$ as positive pairs and all $(\hv_i, \hv_k)$ as negative pairs where $i \ne k$. By maximizing the agreement between positive pairs and minimizing the similarity within negative pairs (i.e. different input instances), in a mini-batch with size $B$, the instance-wise contrastive losses for $h_{\theta}$ and $g_{\theta}$ are
\begin{equation}
    \mathcal{L}_{Inst}^{h}=-\sum_{i=1}^{B}\log \frac{\exp \left(\operatorname{sim}\left(\hv_{i}, \hv_{i}'\right) / \tau\right)}{\sum_{k=1}^{B} \mathbbm{1}_{[k \neq i]} \exp \left(\operatorname{sim}\left(\hv_{i}, \hv_{k}\right) / \tau\right)},
    \label{eq:loss_single1}
\end{equation}
and
\begin{equation}
    \mathcal{L}_{Inst}^{g}=-\sum_{i=1}^{B}\log \frac{\exp \left(\operatorname{sim}\left(\gv_{i}, \gv_{i}'\right) / \tau\right)}{\sum_{k=1}^{B} \mathbbm{1}_{[k \neq i]} \exp \left(\operatorname{sim}\left(\gv_{i}, \gv_{k}\right) / \tau\right)},
    \label{eq:loss_single2}
\end{equation}
where $\mathbbm{1}_{[k \neq i]}$ is the indicator function and $\tau$ is the temperature parameter, $\operatorname{sim}(\cdot, \cdot)$ is the dot product between two $\ell_2$-normalized vectors.

\subsection{Multi-View Contrastive Learning via Co-training}

When the two views $f_{\theta}(\xv)$ and $g_{\theta}(\xv)$ are established, we conduct co-training over them to learn the final representation for $\xv$. We explore two settings: unsupervised and semi-supervised, according to whether the downstream labels are available.
\subsubsection{Why Prototype-based Co-training?} \hfill\\
For contrastive learning, there exist some discussions about the importance of macro semantic structure of the embedding space \cite{haochen2021provable,wang2021pico}. For example, in \cite{haochen2021provable}, they propose to design spectral clustering based contrastive learning loss, which learns the clustering structure in the augmentation space supported by a provable accuracy guarantee. \eat{From an expectation maximization (EM) perspective, the 
theoretical analysis in \cite{wang2021pico} shows that minimizing the distance between the representation and the prototypes leads to a strong intra-class concentration on the representation hypersphere. Thus, in this work, we have the basic assumption that the underlying semantic structure of hidden space is multimodal, where the prototype denotes the center of each component.}However, in multi-view representation learning, there are still several challenges when maintaining these clustering-based semantic structures:
\begin{itemize}
\setlength{\itemsep}{1pt}
\setlength{\parsep}{0pt}
\setlength{\parskip}{0pt}
    \item \textbf{Efficiency}: Maintaining such a semantic structure is memory insufficient and costly. 
    \item \textbf{Robustness}: Instance-wise cross-view alignment may be difficult to achieve in practice because of the complementary and disagreed information of different views. 
\end{itemize}

In this work, we propose to use prototypes to capture similar macro semantic structures between different views. If the view $h$ and $g$ are sufficiently representative, an instance belonging to which prototype should be the same in two views. Here, we define that an instance belongs to a certain prototype if the distance from the instance to that prototype is the closest among all prototypes.
Concretely, we adopt the prototypes for describing the semantic structures of different views. On the one hand, as the discussions mentioned before, prototypes are representative tools for discovering the structures of hidden space. On the other hand, prototypes could provide global information for each instance within the mini-batch, alleviating the dependence on the batch size. Meanwhile, low additional computational and memory costs make the algorithm more efficient. Additionally, by incorporating complementary information and discarding superfluous  information of different views, we design a co-training based representation framework to achieve effective semantic structure alignment. Cross-view prototypes are proposed to share macro information effectively.

\subsubsection{Unsupervised Representation Learning}\hfill\\
\label{sec:unsuper}

We now introduce the prototype-based co-training method proposed for the unsupervised representation learning setting. In this work, we first initiate the intra-view prototypes for each view using K-means clustering, where the clustering centroids denote the prototypes. For the clustering with $C$ clusters, we have the prototypes for view $h$ (resp. view $g$) by
\begin{equation}
    \left\{\uv^h_{1},...,\uv^h_{C}\right\} = \operatorname{K-Means}\left(\left\{\hv_{1},...,\hv_{N}\right\}\right). 
    \label{eq:kmeans1}
\end{equation}

We then calculate the clustering index for $\xv_i$ by assigning it to its nearest prototype in view $h$ (resp. view $g$),
\begin{equation}
    s^h_{i} = \operatorname{argmin}_{j\in \{1, \ldots, C\}}~ \| \hv_{i}-\uv^h_{j} \|.
    \label{eq:ass_1}
\end{equation}
Regardless of the cluster indexes of the two views, we expect $s^h_{i}$ and $s^g_{i}$ should be the same, i.e., the view $h$ and $g$ should share the same clustering structure. Here we use co-training to achieve this goal. Firstly, we use the clustering indexes from the other view to re-calculate the prototype, which we call cross-view prototype. Formally, we have
\begin{equation}
    \tilde{\uv}^h_{c}=\dfrac{\sum_{i=1}^{N}\mathbbm{1}{\left(s^g_{i}=c\right)} \cdot \hv_{i}}{\sum_{i=1}^{N}\mathbbm{1}{\left(s^g_{i}=c\right)}}
    \label{eq:cross1}
\end{equation}
and
\begin{equation}
    \tilde{\uv}^g_{c}=\dfrac{\sum_{i=1}^{N}\mathbbm{1}{\left(s^h_{i}=c\right)} \cdot \gv_{i}}{\sum_{i=1}^{N}\mathbbm{1}{\left(s^h_{i}=c\right)}},
    \label{eq:cross2}
\end{equation}
where $c \in \{1, \ldots, C\}$. 

Based on the cross-view prototypes defined in Equation (\ref{eq:cross1}), similar to \cite{li2020prototypical}, the prototype-based contrastive co-training losses are given by:
\begin{equation}
    \mathcal{L}_{CoT}^{h}=-\sum_{i=1}^{B}\operatorname{log}\dfrac{\operatorname{exp}\left(\operatorname{sim}(\boldsymbol{h}_{i},\tilde{\boldsymbol{q}}^h_{i})\right)}{\sum_{j=1}^{C}\operatorname{exp}\left(\operatorname{sim}(\boldsymbol{h}_{i},\tilde{\boldsymbol{\mu}}^h_{j})\right)},
     \label{eq:loss_cross1}
\end{equation}
and
\begin{equation}
    \mathcal{L}_{CoT}^{g}=-\sum_{i=1}^{B}\operatorname{log}\dfrac{\operatorname{exp}\left(\operatorname{sim}(\boldsymbol{g}_{i},\tilde{\boldsymbol{q}}^g_{i})\right)}{\sum_{j=1}^{C}\operatorname{exp}\left(\operatorname{sim}(\boldsymbol{g}_{i},\tilde{\boldsymbol{\mu}}^g_{j})\right)},
    \label{eq:loss_cross2}
\end{equation}
where $\tilde{\boldsymbol{q}}^h_{i}$ and $\tilde{\boldsymbol{q}}^g_{i}$ are the corresponding cross-view prototype of instance $\xv_i$, i.e,
\begin{equation}
    \tilde{\boldsymbol{q}}^h_{i} = \tilde{\boldsymbol{\mu}}^h_{c}, ~~ \textrm{where}~~c={s}^g_{i},\label{eq:cal_q1}
\end{equation}
and
\begin{equation}
    \tilde{\boldsymbol{q}}^g_{i} = \tilde{\boldsymbol{\mu}}^g_{c}, ~~ \textrm{where}~~c={s}^h_{i}. \label{eq:cal_q}
\end{equation}

Meanwhile, in light of the computational cost and instability of the clustering algorithm, we update the intra-view prototypes in a moving-average way. To be specific, the updated prototype is initialized by $\uv_{c}^{\prime h}=\tilde{\uv}_{c}^h$\eat{$\tilde{\uv}_{c}^h$}. While learning the representations, we update each prototype with the corresponding instances by 
\eat{\begin{equation}
    \uv_{c}^{\prime h} = (1-\gamma) \uv_{c}^{\prime h} + \gamma \bar{\boldsymbol{h}}_c, ~\textrm{where}~      \bar{\boldsymbol{h}}_{c} = \dfrac{\sum_{i=1}^{B}\mathbbm{1}{\left(\tilde{s}^h_{i}=c\right)}\boldsymbol{h}_{i}}{\sum_{i=1}^{B}\mathbbm{1}{\left(\tilde{s}^h_{i}=c\right)}}.
    \label{eq:update1}
\end{equation}}
\begin{equation}
    \uv_{c}^{\prime h} = (1-\gamma) \cdot \uv_{c}^{\prime h} + \gamma \cdot \bar{\boldsymbol{h}}_c,
    \label{eq:update1}
\end{equation}
where 
\begin{equation}
    \bar{\hv}_{c} = \sum_{i=1}^{B}\mathbbm{1}{\left({s}^g_{i}=c\right)} \cdot \hv_{i} / \sum_{i=1}^{B}\mathbbm{1}{\left({s}^g_{i}=c\right)}.
\end{equation} Similarly, we could obtain the updated prototypes of view $g$. The updated prototypes $\uv_{c}^{\prime h}$ and $\uv_{c}^{\prime g}$ will serve as the intra-view prototypes in the next training epoch instead of using the result of clustering algorithm in Eq.(\ref{eq:kmeans1}).

\subsubsection{Semi-Supervised Representation Learning}\hfill\\
\label{sec:semi_super}Based on the co-training paradigm, we are able to extend our proposed model to semi-supervised representation learning easily. In this scenario, we have limited labels (with size $N'$) while performing representation learning. We hope to engage the available labels in the pre-training representation learning phase, rather than just fine-tuning the model prior to downstream classification.

The model in the semi-supervised setting is similar to the one in the unsupervised case. The main difference is that we utilize the mean of the labeled embedding for each ground truth class as the intra-view prototypes defined in Equation (\ref{eq:kmeans1}): 
\begin{equation}
    \boldsymbol{\mu}^h_{c} = \dfrac{\sum_{i=1}^{N'}\mathbbm{1}{\left(Y_{i}=c\right)} \cdot \boldsymbol{h}_{i}}{\sum_{i=1}^{N'}\mathbbm{1}{\left(Y_{i}=c\right)}}, 
    \label{eq:semi_means1}
\end{equation}
where $Y_{i}$ is the ground-truth label of sample $i$. The semi-supervised prototypes $\boldsymbol{\mu}^g_{c} $ could be obtained similarly for view $g$.

After getting the prototypes $\boldsymbol{\mu}^h_{c}$ and $\boldsymbol{\mu}^g_{c}$, the semi-supervised algorithm shares the same implementation as the unsupervised one.

\begin{algorithm}[tb]
	\caption{Framework of Unsupervised Prototype-based Co-Training}
	\label{alg:algorithm1}
  \KwIn{Training dataset $\mathcal{X}$, co-training encoding network $h_\theta, g_\theta$, number of prototypes $C$.}
  
   \KwOut{$h_\theta, g_\theta$}
  \For{$\text{epoch}=0,1,2,...$}{

    Compute the prototypes and prototype assignments of different views $\boldsymbol{\mu^{h}, \mu^{g}}, s_i^h,s_i^g$ with Eq.(\ref{eq:kmeans1})(\ref{eq:ass_1}) or Eq.(\ref{eq:update1}).

     Compute the cross-view prototype for each instance $\tilde{\boldsymbol{q}}^h,\tilde{\boldsymbol{q}}^g$  with Eq.(\ref{eq:ass_1}) to Eq.(\ref{eq:cross2}) and Eq.(\ref{eq:cal_q1})(\ref{eq:cal_q}).

    \For{iter=0,1,2,...}{
     Sample a mini-batch $B$ from $\mathcal{X}$.
  
    \For{$\boldsymbol{x}_{i} \in B$}{
     Obtain the representation:
     $\hv_i=h_{\theta}(\xv_i),\hv'_{i}=\operatorname{Dropout}(h_{\theta}(\xv_{i}))$
     
    $\gv_i=g_{\theta}(\xv_i),\gv'_{i}=\operatorname{Dropout}(g_{\theta}(\xv_{i}))$

       Calculate $\mathcal{L}_{Inst}^{g}, \mathcal{L}_{Inst}^{h}$ with Eq.(\ref{eq:loss_single1})(\ref{eq:loss_single2}).
      
       Calculate $\mathcal{L}^{g}_{CoT}, \mathcal{L}^{h}_{CoT}$ with Eq.(\ref{eq:loss_cross1})(\ref{eq:loss_cross2}).
       
       Update the prototypes in a moving-average style with Eq.(\ref{eq:update1}).
    }
    
    $\mathcal{L}=\mathcal{L}_{Inst}^{g}+\mathcal{L}_{Inst}^{h}+\lambda(  \mathcal{L}_{CoT}^{g}+\mathcal{L}_{CoT}^{h})$
    
    Update $h_\theta, g_\theta$ with $\mathcal{L}$.
    }
    }


\end{algorithm}

\subsubsection{Co-training Algorithm}\hfill\\
The overall loss function of our proposed model is given by:
\begin{equation}
    \mathcal{L}=\mathcal{L}_{Inst}^{g}+\mathcal{L}_{Inst}^{h}+\lambda(  \mathcal{L}_{CoT}^{g}+\mathcal{L}_{CoT}^{h})
    \label{eq:loss}
\end{equation}
where $\lambda $ is a hyper-parameter to balance the contrastive loss of single-view encoders and multi-view co-training modules. Take the unsupervised prototype-based co-training algorithm as an example, the complete algorithm is shown in \textbf{Algorithm \ref{alg:algorithm1}}.

For implementation, we perform the calculation of prototypes at the beginning of each epoch. For obtaining two different views of the same time series, we utilize the original time series and the frequency signals from fast fourier transform (FFT) to be the input of $h_\theta, g_\theta$, respectively. However, we do have other choices for constructing different views. For example, use the cropped series \cite{yue2022ts2vec}, utilize the seasonal-trend decomposition \cite{cleveland90}, find out the pre-defined shaplets \cite{ye2009time,li2022ips}, or transform the original signals with discrete wavelet \cite{zhou2022fedformer}. Hopefully, our proposed model is a general framework for dealing with multi-view time series data.

\section{Experiments}

\begin{table}[tb]
\vspace{-3mm}
\caption{Statistical summary of experimental datasets.}
\vspace{-3mm}
\resizebox{\columnwidth}{!}{%
\begin{tabular}{cccccc}
\toprule
         & \textbf{HAR}  &\textbf{ Sleep-EDF} &\textbf{ Epilepsy} & \textbf{Waveform}  & \textbf{Gesture} \\
\midrule
\textbf{\# Train }& 7352 & 35503     & 9200     & 59922 & 320   \\
\textbf{\# Test}  & 2947 & 6805      & 2300     & 16645  & 120  \\
\textbf{Length }  & 128  & 3000      & 178      & 2500   &315  \\
\textbf{Channel}  & 9    & 1         & 1        & 2   & 3     \\
\textbf{\# Class} & 6    & 5         & 2        & 4   & 8    \\
\bottomrule
\end{tabular}}
\label{table:dataset}
\vspace{-3mm}
\end{table}

\begin{table*}[htb]
\centering
\caption{Experiment results (mean performance and standard deviation) in unsupervised representation learning for time series classification. The accuracy (\%) and AUROC (\%) of the linear classifier are reported.}
\resizebox{0.93\textwidth}{!}{%
\begin{tabular}{ccccccccc}
\toprule
\multicolumn{1}{c}{\multirow{2}{*}{\textbf{Methods}}}  &
  \multicolumn{2}{c}{\textbf{HAR}} &
  \multicolumn{2}{c}{\textbf{Sleep-EDF}} &
  \multicolumn{2}{c}{\textbf{Epilepsy}} &
  \multicolumn{2}{c}{\textbf{Waveform}} \\
  \cmidrule(l{10pt}r{10pt}){2-3}\cmidrule(l{10pt}r{10pt}){4-5}
  \cmidrule(l{10pt}r{10pt}){6-7}\cmidrule(l{10pt}r{10pt}){8-9}
               \multicolumn{1}{c}{}    & \textbf{Accuracy} & \textbf{AUROC} & \textbf{Accuracy} & \textbf{AUROC} & \textbf{Accuracy} & \textbf{AUROC} & \textbf{Accuracy} & \textbf{AUROC} \\ \midrule

\textbf{TNC}       & 80.20$\pm$0.61     & 71.58$\pm$0.93 & 52.86$\pm$0.43    & 55.36$\pm$0.37 & 81.26$\pm$0.53    & 83.56$\pm$1.29 & 59.50$\pm$1.98     & 55.10$\pm$2.86  \\
\textbf{SimCLR}    & 86.53$\pm$0.53    & 87.98$\pm$0.14 & 65.57$\pm$0.11    & 75.98$\pm$0.64 & 96.03$\pm$0.22    & 98.82$\pm$0.11 & 65.23$\pm$1.91    & 64.65$\pm$3.27 \\
\textbf{Mixing-Up} & 90.46$\pm$0.18    & 91.92$\pm$0.78 & 72.16$\pm$0.14    & 74.68$\pm$0.07 & 96.49$\pm$0.58    & 99.10$\pm$0.10   & 52.46$\pm$0.58    & 49.42$\pm$0.53 \\

\textbf{CoST}      & 69.26$\pm$1.57    & 80.88$\pm$0.05 & 69.87$\pm$3.63    & 69.54$\pm$0.41 & 94.41$\pm$0.11    & 98.67$\pm$0.01 & 53.68$\pm$0.30     & 56.47$\pm$0.01 \\
\textbf{TF-C}      & 90.22$\pm$0.70    & 88.41$\pm$0.14 & 64.27$\pm$0.24    & 75.82$\pm$0.28 & 96.75$\pm$0.36     & 98.56$\pm$0.36 & 79.72$\pm$0.30    & 60.58$\pm$0.11 \\
\textbf{TS-TCC}    & 90.57$\pm$0.20     & 94.70$\pm$0.04  & 80.50$\pm$0.05     & 80.02$\pm$0.18 & 97.29$\pm$0.06    & 98.66$\pm$0.64 & 65.43$\pm$2.11    & 58.03$\pm$0.04 \\
\textbf{TS2Vec}    & 92.62$\pm$0.05    & 96.96$\pm$0.32 & 73.56$\pm$3.51    & 91.78$\pm$1.73 & 97.32$\pm$0.12    & 99.45$\pm$0.01 & 73.09$\pm$2.75    & 56.83$\pm$2.67 \\
\textbf{TS-CoT} &
  \textbf{94.05$\pm$0.24} &
  \textbf{99.63$\pm$0.23} &
  \textbf{80.99$\pm$0.27} &
  \textbf{95.16$\pm$0.13} &
  \textbf{98.10$\pm$0.23} &
  \textbf{99.69$\pm$0.06} &
  \textbf{83.06$\pm$1.23} &
  \textbf{80.44$\pm$0.21} \\ \bottomrule
\end{tabular}
}
\label{table:unsuper}
\end{table*}

In this section, we conduct extensive empirical evaluations and demonstrate the performance of our proposed model on three different tasks, including unsupervised time series representation learning, semi-supervised time series representation learning, and self-supervised time series pre-training.

\subsection{Datasets}
Five real-world time-series datasets will be utilized for evaluation, 

whose statistical summary is shown in Table \ref{table:dataset}. 

\begin{itemize}
    \item \textbf{Human Activity Recognition (HAR)}: The HAR dataset \cite{anguita2013public} contains multi-channel sensor signals of 30 subjects, each performing one out of six possible activities. 3-axial linear acceleration and 3-axial angular velocity are captured at a constant rate of 50Hz. The corresponding experiments contain 30 volunteers within an age bracket of 19-48 years.
    \item \textbf{Sleep-EDF}: The Sleep-EDF dataset \cite{goldberger2000physiobank} includes single-channel EEG signals sampled at 100Hz. The sleep recordings cover five different sleep stages, namely, Wake (W), Non-rapid eye movement (N1, N2, N3), and Rapid Eye Movement (REM). 
    \item \textbf{Epilepsy}: Epilepsy is an epileptic seizure recognition dataset. It includes EEG signals from 500 subjects. We follow the preprocessing of \cite{eldele2021time} for the classification task, which divides the whole datasets into two different classes.
    \item \textbf{Waveform}: ECG Waveform \cite{goldberger2000physiobank, moody1983new} includes the long-term ECG signals of patients with atrial fibrillation. The ECG recordings have two channels with four different classes. All the signals are sampled at 250Hz.
    \item 
\textbf{Gesture}: Gesture \cite{liu2009uwave} dataset contains three-channel accelerometer measurements of the hand movement paths. 440 samples collected at 100Hz cover eight hand gestures.

\end{itemize}

\subsection{Baseline Methods and Implementation}
For unsupervised and semi-supervised classification tasks, we compare our model with the following eight baselines with their public code:

\begin{itemize}

    \item TNC \cite{tonekaboni2020unsupervised} utilizes the local smoothness of a signal's generative process and learns generalizable representations for non-stationary time series. \footnote{https://github.com/sanatonek/TNC\_representation\_learning}
    \item SimCLR \cite{tang2020exploring} is the adoption and adaptation of  SimCLR framework, a contrastive learning technique for visual representations \cite{chen2020simple}. \footnote{https://github.com/iantangc/ContrastiveLearningHAR}
    \item Mixing-Up \cite{wickstrom2022mixing} utilizes the label smoothing strategy and proposes a novel data augmentation scheme by mixing two data samples. \footnote{https://github.com/Wickstrom/MixupContrastiveLearning}
    \item TS-TCC \cite{eldele2021time} combines both weak and strong augmentations for contrastive learning. A prediction based temporal contrasting module is also proposed.
    \footnote{https://github.com/emadeldeen24/TS-TCC}
    \item CoST \cite{woo2022cost} is a paradigm learning disentangled feature representations from a causal perspective. 
    COST is proposed for time series forecasting. 
    We adapt its representation for classification tasks here. 
    \footnote{https://github.com/salesforce/CoST}
    \item TF-C \cite{zhang2022self} is a time series pre-training framework by combining temporal domain embedding and frequency domain embedding. We also adapt TF-C to learn latent representation for classification tasks here. 
    \footnote{https://github.com/mims-harvard/TFC-pretraining}
    \item TS2Vec \cite{yue2022ts2vec} learns representations of time series in an arbitrary semantic level. Multi-level of representations are learned for contextual representations.
    \footnote{https://github.com/yuezhihan/ts2vec}

\end{itemize}

Among the compared baselines, TF-C and CoST involve both temporal and spectral embedding in their models. TF-C treats the frequency signal of the original series as part of the input, and CoST conducts fourier transform on the embedding of each hidden layer. 

For TS-CoT, we use a 3-level CNN-based pyramid network to be single-view encoder with max-pooling layer as the readout function. We concatenate the embedding of different views as the final representation. The representation dimensions of the three tasks are 512, 512, and 128. We set the batch size to 256. We adopt an Adam optimizer with a learning rate 0.001. For the unsupervised and semi-supervised tasks, we set the training epoch to 30, 20, 40, 20 for four experimental datasets HAR, Sleep-EDF, Epilepsy, Waveform. For pre-training task, the training epoch is 50. We first train the model with only single-view contrastive loss in the first several epochs as a warming-up.

\begin{table*}[htb]
\centering
\caption{Experiment results (mean performance and standard deviation) in semi-supervised representation learning (with 10\% labeled data) for time series classification. The accuracy (\%) and AUROC (\%) are reported.}
\resizebox{0.93\textwidth}{!}{

\begin{tabular}{ccccccccc}
\toprule
\multirow{2}{*}{\textbf{Methods}} &
  \multicolumn{2}{c}{\textbf{HAR}} &
  \multicolumn{2}{c}{\textbf{Sleep-EDF}} &
  \multicolumn{2}{c}{\textbf{Epilepsy}} &
  \multicolumn{2}{c}{\textbf{Waveform}} \\
    \cmidrule(l{10pt}r{10pt}){2-3}\cmidrule(l{10pt}r{10pt}){4-5}
  \cmidrule(l{10pt}r{10pt}){6-7}\cmidrule(l{10pt}r{10pt}){8-9}
                   & \textbf{Accuracy} & \textbf{AUROC} & \textbf{Accuracy} & \textbf{AUROC} & \textbf{Accuracy} & \textbf{AUROC} & \textbf{Accuracy} & \textbf{AUROC} \\ 
                   \midrule
\textbf{TNC}       & 82.38$\pm$3.65 & 81.00$\pm$6.83    & 51.34$\pm$11.15 & 62.26$\pm$11.95 & 81.98$\pm$5.19 & 84.98$\pm$4.99 & 63.92$\pm$2.58 & 57.62$\pm$3.02    \\
\textbf{SimCLR}    & 91.38$\pm$0.16 & 98.29$\pm$0.19 & 72.06$\pm$0.39  & 70.47$\pm$0.15  & 96.43$\pm$0.06 & 99.20$\pm$0.47  & 86.66$\pm$3.31 & 70.30$\pm$4.59  \\
\textbf{Mixing-Up} & 91.42$\pm$0.29 & 98.69$\pm$0.07 & 80.16$\pm$1.11  & 94.44$\pm$0.17  & 96.07$\pm$0.20  & 98.31$\pm$0.03 & 60.39$\pm$1.61 & 53.92$\pm$3.85 \\

\textbf{CoST}      & 86.44$\pm$2.76 & 91.42$\pm$1.91 & 65.33$\pm$0.77  & 67.32$\pm$2.97  & 95.06$\pm$1.44 & 97.35$\pm$1.13 & 65.41$\pm$4.52 & 56.39$\pm$3.98 \\
\textbf{TF-C}      & 86.72$\pm$2.03 & 87.49$\pm$0.08 & 64.53$\pm$1.89  & 74.37$\pm$2.27  & 94.53$\pm$0.70  & 96.94$\pm$0.19 & 78.75$\pm$3.02 & 62.84$\pm$3.14 \\
\textbf{TS-TCC}    & 92.23$\pm$0.29 & 96.66$\pm$0.10     & 78.54$\pm$1.20   & 79.41$\pm$0.09  & 96.22$\pm$1.86 & 96.30$\pm$0.34  & 71.05$\pm$2.79 & 59.68$\pm$1.69 \\
\textbf{TS2Vec}    & 90.38$\pm$0.08 & 98.51$\pm$0.36 & 77.46$\pm$0.85  & 93.98$\pm$0.18  & 92.28$\pm$1.99 & 98.37$\pm$0.48 & 83.56$\pm$5.67 & 72.84$\pm$3.14 \\
\textbf{TS-CoT} &
  \textbf{94.27$\pm$0.60} &
  \textbf{99.71$\pm$0.03} &
  \textbf{81.91$\pm$0.21} &
  \textbf{95.36$\pm$0.14} &
  \textbf{97.62$\pm$0.12} &
  \textbf{99.53$\pm$0.04} &
  \textbf{91.48$\pm$0.86} &
  \textbf{81.81$\pm$1.71}\\ \bottomrule
\end{tabular}}
\label{table:semi_super}
\end{table*}

\begin{figure*}[htb]
  \centering
  \includegraphics[width=\textwidth]{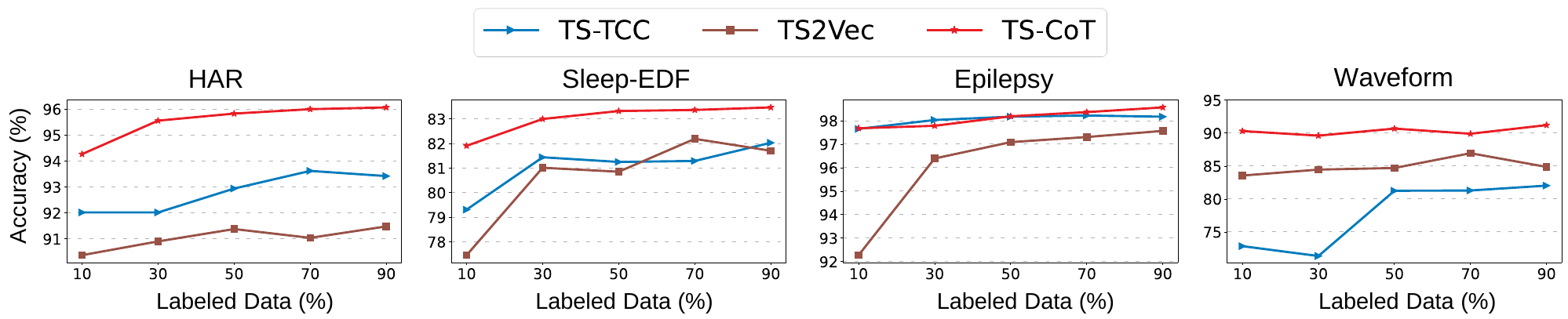}
  \vspace{-8mm}
  \caption{Performance comparison for semi-supervised representation learning with different percentages of labeled data. }
  \label{figure:semi_ratio}
  \vspace{-2mm}
\end{figure*}

\subsection{Performance Comparison}
\subsubsection{Unsupervised Representation Learning}\hfill\\
\textbf{Setup.} In this task, we perform time series classification based on the results of unsupervised representation learning on four datasets.  We train our proposed model without any supervision based on the framework introduced in Section \ref{sec:unsuper}. We perform a multi-way prototype learning with different numbers of prototypes in parallel to describe both coarse and fine-grained structures of the hidden space. Then, we freeze the parameters of the encoder and generate the embeddings for all the samples. For classification evaluation, we train a linear classifier based on the training data. For linear classifier selection, we randomly split 20\% of the training data as a validation set. After running for five trials, we report the mean performance and the standard deviation on the test set. Two metrics, accuracy and the area under the receiver operating characteristic curve (AUROC), are used for performance comparison.

\textbf{Results.}
Table \ref{table:unsuper} shows the results. We find that TS-CoT achieves the best results compared to the baseline representation learning approaches in all benchmarks. On average, TS-CoT outperforms other methods by a large margin. Concretely, our TS-CoT gains more than 10\% accuracy improvement on \textbf{Waveform} dataset, which is the largest one among all the experimental datasets. It indicates the superiority of our model in dealing with large-scale datasets, which could be attributed to the macro structure awareness in the representation space by using the prototype-based co-training strategy. Meanwhile, compared with the existing spectral-temporal approach (i.e. CoST and TF-C), TS-CoT is able to learn better representations by utilizing the complementary information of different views. The superiority over the instance-wise cross-view alignment can be observed.

 \begin{figure*}[htb]
  \centering
  \includegraphics[width=0.95\textwidth]{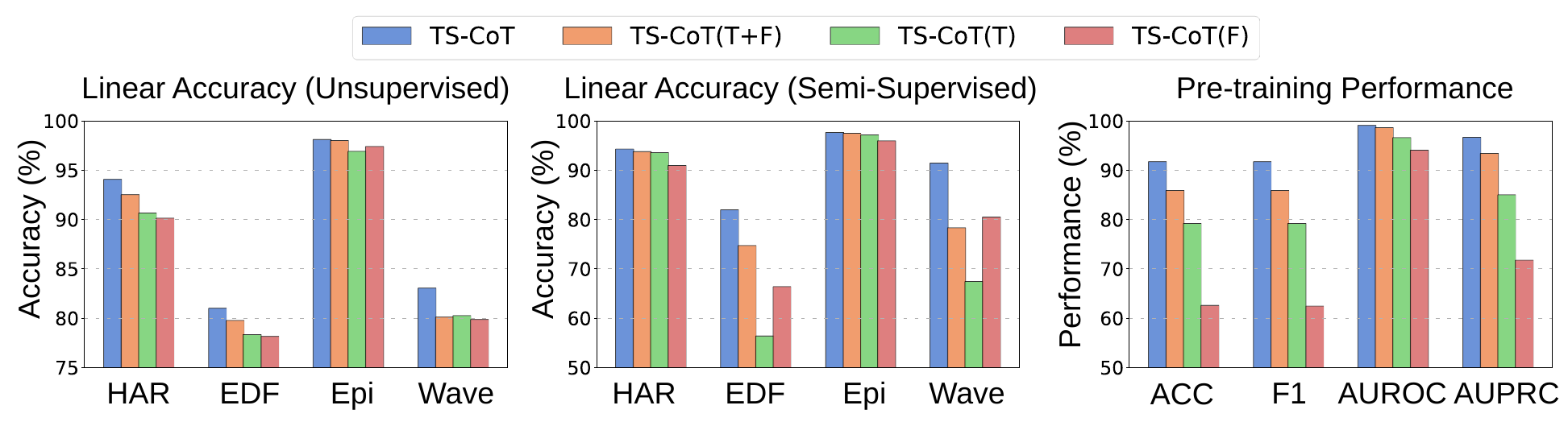}
  \vspace{-3mm}
  \caption{Ablation study. The comparison of TS-CoT to different ablation variants on three experimental tasks. We abbreviate "Sleep-EDF", "Epilepsy" and "Waveform" to "EDF", "Epi" and "Wave", respectively. }
  \label{figure:ablation}
\end{figure*}

\vspace{-3mm}
\subsubsection{Semi-supervised Representation Learning} \hfill\\
\textbf{Setup.} In the semi-supervised setting, we investigate the scenario in that we have only a fraction of labeled data (10\%) while training the model. For our TS-CoT, with accessible labels, we train the representation learning model as described in Section \ref{sec:semi_super}. For other competing methods, the model is trained in an unsupervised way. Then we fine-tune the baseline models by using the labeled data. We have a similar train-validation split to the unsupervised task. Mean performance and standard deviation after 5 runs are reported.

\textbf{Results.}
The results are shown in Table \ref{table:semi_super}. The superior performance of our proposed TS-CoT could be observed clearly. Considering that only a small fraction of labels are accessible for fine-tuning, it is easy to be overfitted for other baseline methods. This issue may be alleviated in our proposed TS-CoT, because we employ the limited labels not only for fine-tuning but also for representation model training. 

The efficiency of TS-CoT could be further illustrated in Figure \ref{figure:semi_ratio}. We compare TS-CoT with two competing methods TS-TCC and TS2Vec with available label ratios varying from 10\% to 90\%. TS-CoT outperforms others on 3 out of 4 datasets and has competitive results on \textbf{Epilepsy}. It should be noted that TS-CoT has similar classification performance with different label ratios. TS-TCC, as well as TS2Vec, perform poorly when the label size becomes small. It substantiates that TS-CoT could take advantage of the limited labels more efficiently and alleviate the overfitting by involving the labels in both representation learning and model fine-tuning. Our proposed TS-CoT effectively harnesses the limited available supervision.
\vspace{-1mm}
\subsubsection{Self-Supervised Time Series Pre-training}\hfill\\
\textbf{Setup.}
In this part, we follow the one-to-one evaluation setting of \cite{zhang2022self}. We pre-train the representation learning model on one pre-training dataset HAR (recording 6 different daily activities) and fine-tune it on one target dataset Gesture (describing 8 hand gestures). Acceleration signals are recorded in both the pre-training and the target dataset. For varying length and dimensionality of time series in pre-training and target datasets, referring to the setup of \cite{zhang2022self}, we apply our model to one channel and make different channels share the encoder.
For performance comparison, five metrics (Accuracy, Precision, Recall, F1-Score, AUROC) are listed. 

\textbf{Results.}
Mean and standard deviation of performance after 5 independent trials are shown in Table \ref{table:pretrain}. We report some baseline results from \cite{zhang2022self} with the same experimental setup. We observe that TS-CoT shows superiority over other baselines including TF-C, a time-frequency consistency learning model designed for pre-training. The performance improvement also proves the importance of the co-training strategy of TS-CoT.

\begin{table}[tb]
\centering
\caption{Performance comparison (\%) of One-to-One pre-training evaluation. The results of baselines with * are reported by \cite{zhang2022self}. }
\vspace{-2mm}
\resizebox{\columnwidth}{!}{
\begin{tabular}{c|ccccc}
\toprule
                & \textbf{Accuracy} & \textbf{Precision} & \textbf{Recall} & \textbf{F1}    & \textbf{AUROC} \\ \midrule
\textbf{TNC}    & 51.32$\pm$0.31    & 50.84$\pm$0.78     & 51.32$\pm$0.31   & 51.27$\pm$0.21 & 83.70$\pm$0.13  \\
\textbf{SimCLR*}    & 43.83$\pm$6.52 & 42.44$\pm$10.72 & 43.83$\pm$6.52 & 37.13$\pm$9.19 & 77.21$\pm$5.59  \\
\textbf{Mixing-Up*} & 71.83$\pm$1.23 & 70.01$\pm$1.66  & 71.83$\pm$1.23 & 69.91$\pm$1.45 & 91.27$\pm$0.18\\

\textbf{CoST}   & 83.47$\pm$0.76    & 85.94$\pm$0.20      & 83.47$\pm$0.76  & 83.86$\pm$1.14 & 96.25$\pm$0.03  \\
\textbf{TF-C*}   & 78.24$\pm$2.37    & 79.82$\pm$4.96     & 80.11$\pm$3.22  & 79.91$\pm$2.96 & 90.52$\pm$1.36 \\
\textbf{TS-TCC*} & 75.93$\pm$2.42    & 76.68$\pm$2.57     & 75.66$\pm$2.31  & 74.57$\pm$2.10  & 88.66$\pm$0.40   \\
\textbf{TS2Vec*} & 64.53$\pm$2.60     & 62.87$\pm$3.39     & 64.51$\pm$2.18  & 62.61$\pm$2.94 & 88.90$\pm$0.54  \\
\textbf{TS-CoT}    & \textbf{91.67$\pm$0.75} & \textbf{91.95$\pm$0.85}  & \textbf{91.67$\pm$0.75} & \textbf{91.70$\pm$0.75} & \textbf{99.15$\pm$0.05} \\
\bottomrule
\end{tabular}
}

\label{table:pretrain}
\vspace{-2mm}
\end{table}

\subsection{Effectiveness of Key Components}
To validate the effectiveness of the co-training strategy in TS-CoT, we compare it with several ablations. We establish two ablation variants by utilizing the representation of one single view, denoted by \textit{TS-CoT(T) and TS-CoT(F)} here. Another variant (denoted by \textit{TS-CoT(T+F)}) comes from a simple combination of different views, i.e., by concatenating the representations from two views together. As shown in Figure \ref{figure:ablation}, the full model, TS-CoT, consistently achieves the best performance compared with other variants on all three experimental settings. The effectiveness of the proposed multi-view could be validated. In addition, we can not get a consistent comparison between the single-view variant \textit{TS-CoT(T)} and \textit{TS-CoT(F)}. It illustrates the necessity of considering complementary information from different views in order to obtain a better representation.

\begin{figure}[tb]
  \centering
  \includegraphics[width=0.65\columnwidth]{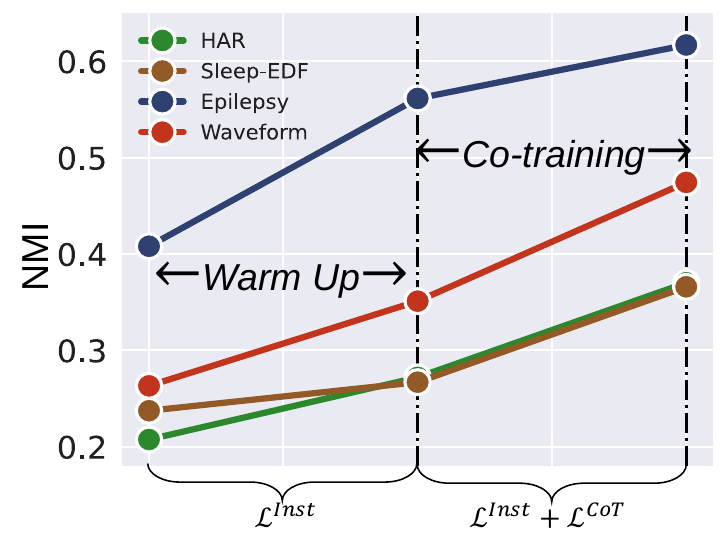}
  \vspace{-3mm}
  \caption{Training progress and NMI of TS-CoT on four experimental datasets.}
  \label{figure:nmi}
  
\end{figure}

In addition, to verify that our proposed model could capture the semantic structure successfully, we compare the representation clustering results of the randomly initialized model, the model trained by single-view warm-up, and the full model trained by multi-view co-training. We show the normalized mutual information (NMI) of unsupervised representation clustering in Figure \ref{figure:nmi}. We can observe that the NMI increases after single-view warm-up training. Next, via our prototype-based multi-view co-training, the semantic clustering structure is enhanced in the hidden space, which further increases the NMI significantly. The observed increase in NMI throughout the TS-CoT training process demonstrates the effective capture of semantic structure by our proposed algorithm.

\subsection{Robustness against Noisy Data}

\begin{figure}[htb]
  \centering
  \includegraphics[width=\columnwidth]{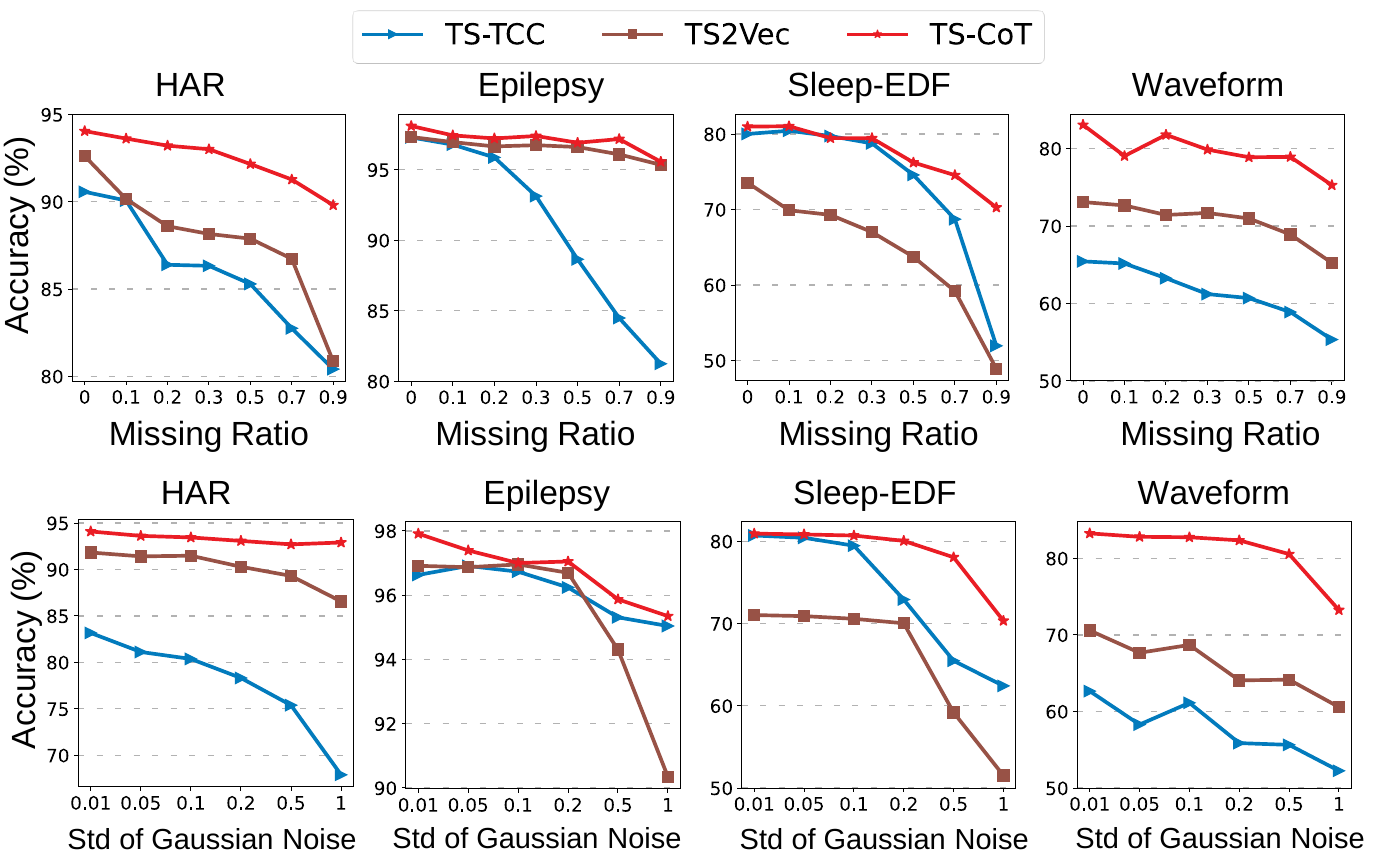}
  \vspace{-7mm}
  \caption{Unsupervised representation learning performance with different missing ratios or different standard deviations of Gaussian noise.} 
  \vspace{-7mm}
  \label{figure:robustness}
  
\end{figure}

In this subsection, we conducted experiments to investigate the robustness of our proposed model against noisy data. We performed unsupervised representation learning experiments on two typical types of data noise, random missingness and Gaussian noise (with zero mean and varying standard deviation). In detail, we compared our proposed TS-CoT with two competitive baselines, namely TS2Vec and TS-TCC. The experimental results, depicted in Figure \ref{figure:robustness}, indicate that although the downstream classification accuracy decreases for all three methods due to noise enlargement, our proposed TS-CoT consistently demonstrates superior robustness for two missing mechanisms across four experimental datasets.

Despite the presence of data noise during model training, we could still effectively leverage the available information through multi-view co-training. This further demonstrates the ability of our proposed method to integrate complementary cross-view information and effectively handle differences among different views.

\vspace{-4mm}

\subsection{Sensitivity Analysis}
\begin{figure}[htb]
  \centering
  \includegraphics[width=\columnwidth]{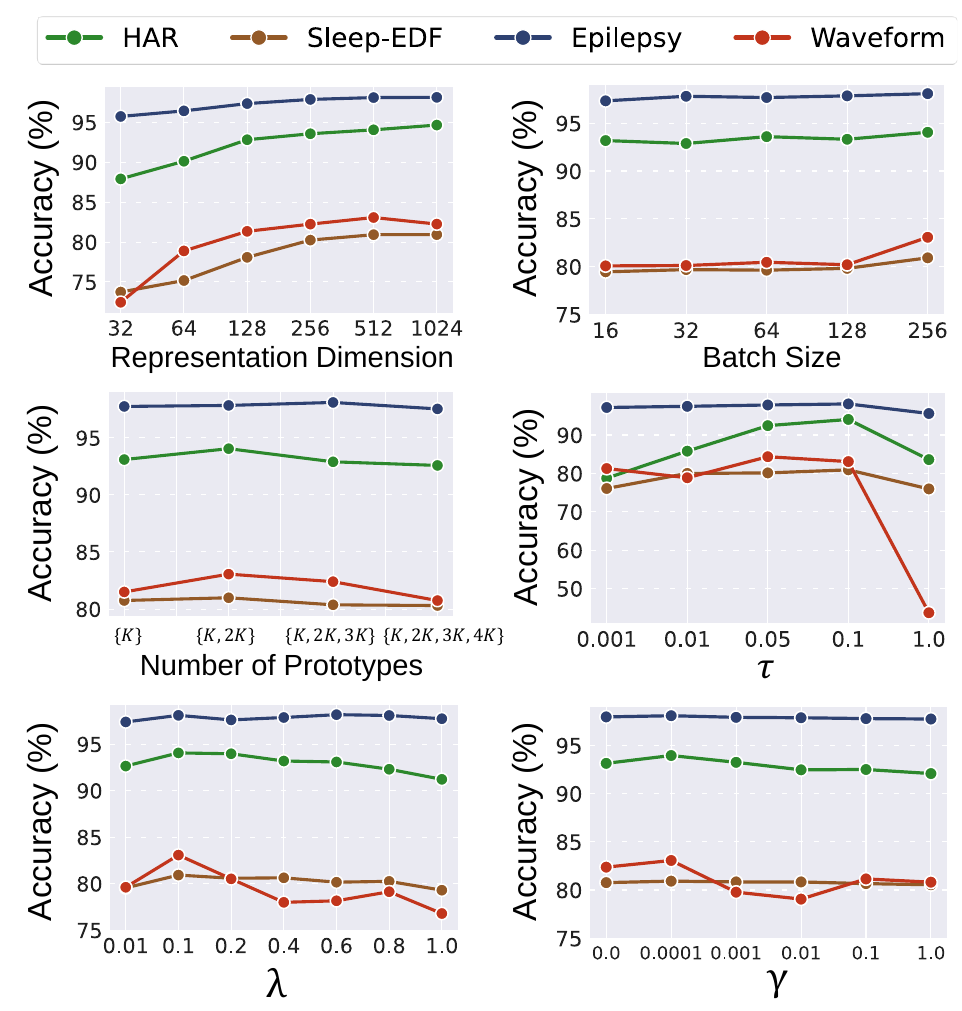}
  \vspace{-7mm}
  \caption{Sensitivity analysis on unsupervised representation learning. The performances of the linear classifier are reported.} 
  \vspace{-5mm}
  \label{figure:sensitive}
  
\end{figure}

We further investigate the sensitivity of TS-CoT to different hyperparameters in the unsupervised representation learning task. Several hyperparameters, representation dimension, batch size, number of prototypes, $\tau$, $\lambda$, and $\gamma$ are analyzed in Figure \ref{figure:sensitive}. The downstream classification accuracy benefits from a large representation dimension, but the improvement becomes marginal when the representation dimension exceeds 256. Moreover, we observe the performance of TS-CoT is robust to different batch sizes. Given that the prototype-based co-training could provide each mini-batch with information on global structure, the limitation of batch size could be alleviated in our model consequently. 

Additionally, our proposed TS-CoT model enables us to perform multi-way prototype learning with varying numbers of prototypes, allowing us to capture both coarse and fine-grained semantic structures in the hidden spaces. Let K denote the number of ground-truth classes. Through the comparison of various prototype numbers, we observed that a moderate prototype setting, specifically a two-way prototype \{K,2K\}, is preferable. Coarse prototype settings may overlook important hidden structure details, while excessively fine-grained settings can impede the quick discovery of the appropriate semantic structure.

 Meanwhile, similar to other contrastive learning based methods, the performance is quite sensitive to the value of temperature $\tau$. Optimal $\tau$ has different values on different datasets. $0.05, 0.1$ are two robust temperatures for most time series data.
 TS-CoT is also robust to the loss balancing parameter $\lambda$ and the moving average parameter $\gamma$. An empirical approach for choosing $\lambda$ is to make two parts of Eq.(\ref{eq:loss}) have nearly balanced values. For $\gamma$ selection, a small value is preferred. Though the moving average parameter $\gamma$ is small, the prototypes could still get enough updates in light of the scale of the training sets.

\vspace{-4mm}

\section{Conclusions and Future Work}
In this work, we present a prototype-based co-training algorithm for noisy time series representation learning and demonstrate its effectiveness through extensive experiments. We shed light on the possibility of combining contrastive learning and co-training for representation learning through prototypes. We investigate the model robustness by exploring complementary information of different views of time series data. Two specific views (time and frequency) are used in this paper, whereas other view construction ways are not fully explored. When and which kind of multi-views to choose will also be an interesting problem, and we leave it to our future work.

\begin{acks}
This research is supported in part by Foshan HKUST Projects FSUST20-FYTRI03B, NSFC Grant No. 62206067, Guangzhou-HKUST(GZ) Joint Funding Scheme 2023A03J0673, and the Industrial Informatics and Intelligence (Triple-i) institute at HKUST(GZ).
\end{acks}

\clearpage
\bibliography{example_paper}

\bibliographystyle{ACM-Reference-Format}

\end{document}